\documentclass[10pt,twocolumn,letterpaper]{article}

\usepackage[pagenumbers]{cvpr} 

\definecolor{cvprblue}{rgb}{0.21,0.49,0.74}
\usepackage[pagebackref,breaklinks,colorlinks,allcolors=cvprblue]{hyperref}

\usepackage{url}
\usepackage[utf8]{inputenc} 
\usepackage[T1]{fontenc}    
\usepackage{hyperref}       
\usepackage{url}            
\usepackage{booktabs}       
\usepackage{amsfonts}       
\usepackage{nicefrac}       
\usepackage{microtype}      
\usepackage{xcolor}         

\usepackage{graphicx}
\usepackage{amsmath}
\usepackage{amssymb}
\usepackage{booktabs}

\usepackage{array}
\usepackage{caption}
\usepackage{multirow}
\usepackage{makecell}
\usepackage{pifont}

\usepackage{colortbl}
\usepackage{enumitem}
\usepackage[normalem]{ulem}
\usepackage{adjustbox}
\usepackage{cellspace}
\usepackage{float}
\usepackage{amssymb}
\usepackage{booktabs}
\usepackage{cvpr}
\usepackage{wrapfig}
\usepackage{algorithm}
\usepackage{algorithmic}
\usepackage{eqparbox}

\newcommand{\tabincell}[2]{\begin{tabular}{@{}#1@{}}#2\end{tabular}}
\newlength\savewidth

\def\ours{\emph{VideoMage}}
\def\sampling{SCS}
\def\paperID{8206} 
\def\confName{CVPR}
\def\confYear{2025}

\makeatletter
\g@addto@macro\@maketitle{
    \begin{figure}[H]
        \setlength{\linewidth}{\textwidth}
        \setlength{\hsize}{\textwidth}
        \centering

        \vspace{-7mm}

        \includegraphics[width=\linewidth]{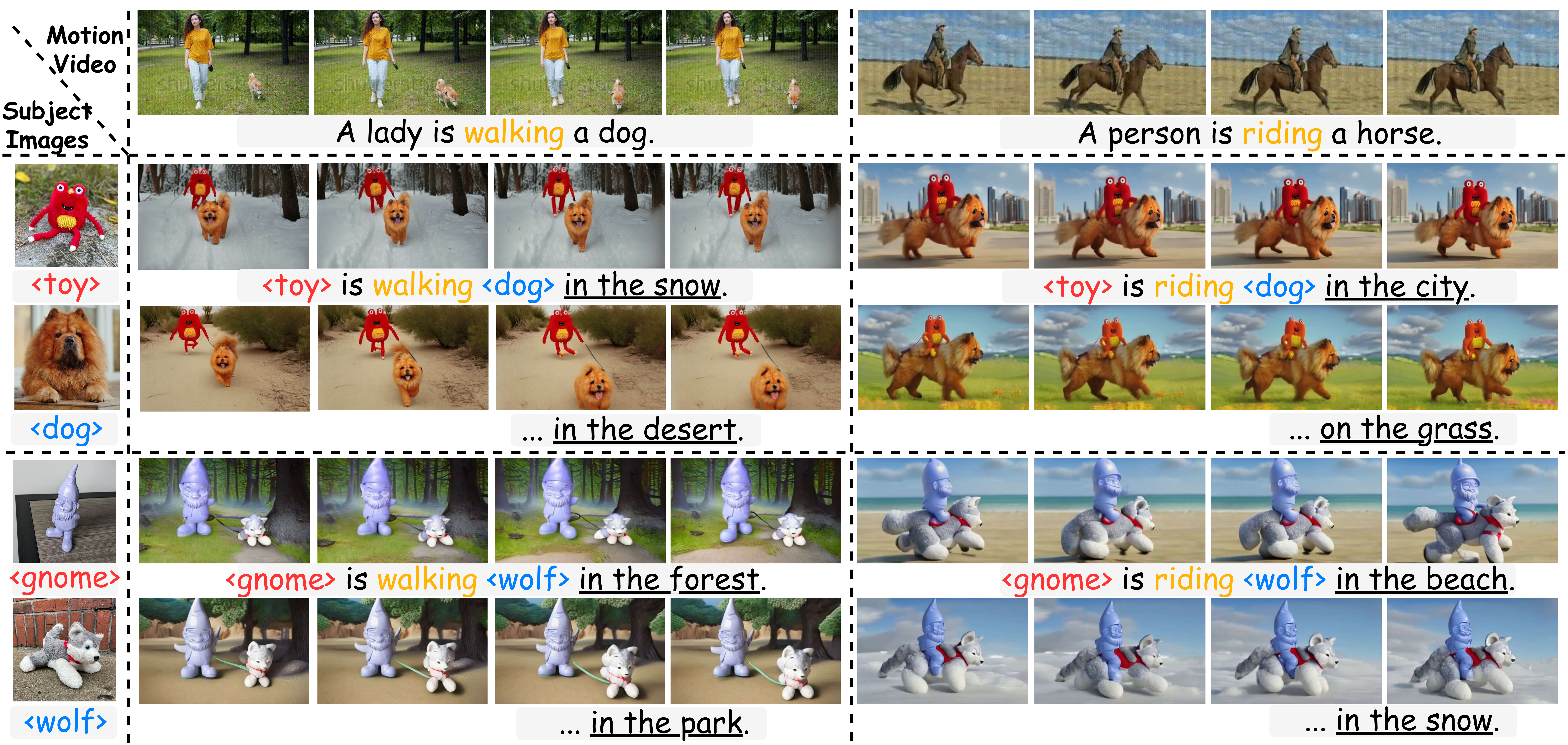}
        \caption{\textbf{Illustration of multi-subject and motion customization.} Given images of multiple subjects, a reference motion video, and a related text prompt, our \ours{} is able to produce video outputs aligned with such inputs.}
        \label{fig:teaser}
    \end{figure}
}
\makeatother

\title{\includegraphics[width=0.05\textwidth]{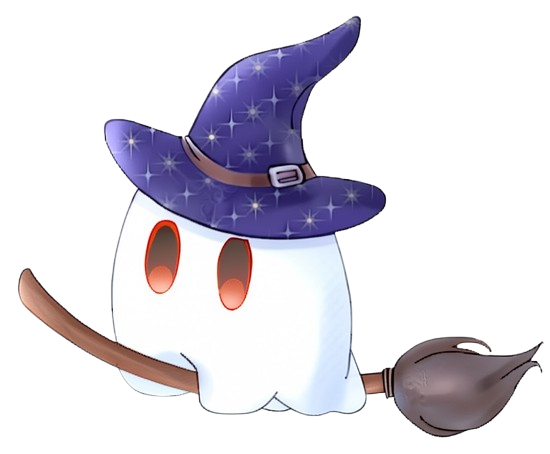}\ours{}: Multi-Subject and Motion Customization\\ of Text-to-Video Diffusion Models}

\author{
    {Chi-Pin Huang}$^{1,\dagger}$, {Yen-Siang Wu}$^2$, {Hung-Kai Chung}$^2$, \\
    {Kai-Po Chang}$^1$, {Fu-En Yang}$^3$, and {Yu-Chiang Frank Wang}$^{1,3,\ddagger}$\\
    \normalsize \textsuperscript{1} Graduate Institute of Communication Engineering, National Taiwan University\\
    \normalsize \textsuperscript{2} National Taiwan University \quad \textsuperscript{3} NVIDIA\\
    {\tt\small $^{\dagger}$f11942097@ntu.edu.tw, $^{\ddagger}$frankwang@nvidia.com}
}

\begin{document}
\maketitle
\begin{abstract}
\label{sec:abstract}

Customized text-to-video generation aims to produce high-quality videos that incorporate user-specified subject identities or motion patterns. However, existing methods mainly focus on personalizing a single concept, either subject identity or motion pattern, limiting their effectiveness for multiple subjects with the desired motion patterns. To tackle this challenge, we propose a unified framework VideoMage for video customization over both multiple subjects and their interactive motions. VideoMage employs subject and motion LoRAs to capture personalized content from user-provided images and videos, along with an appearance-agnostic motion learning approach to disentangle motion patterns from visual appearance. Furthermore, we develop a spatial-temporal composition scheme to guide interactions among subjects within the desired motion patterns. Extensive experiments demonstrate that VideoMage outperforms existing methods, generating coherent, user-controlled videos with consistent subject identities and interactions. Project Page: \textnormal{\href{https://jasper0314-huang.github.io/videomage-customization}{\textcolor{magenta}{https://jasper0314-huang.github.io/videomage-customization}}}

\end{abstract}    
\section{Introduction}
\label{sec:intro}

In recent years, unprecedented success in diffusion models~\cite{ho2020ddpm, song2020ddim, song2020score} has greatly improved the generation of photorealistic videos~\cite{ho2022vdm, ho2022imagenvideo, singer2022makeavideo, blattmann2023align, he2022lvdm} from textual descriptions, enabling new possibilities for video content creation. However, while high-quality and diverse videos can now be synthesized, relying solely on text descriptions cannot offer precise control over desirable content that accurately aligns with user intents~\cite{jiang2024videobooth}. Therefore, customizing user-specific video concepts from the provided references draws significant attention from both academics and industry.

To address this challenge, several works~\cite{jiang2024videobooth, guo2023animatediff, wei2024dreamvideo, wang2024customvideo} have explored customizing a desirable \textit{subject identity} into synthesized videos.
For example, AnimateDiff~\cite{guo2023animatediff} animates the user-provided subject by tuning temporal modules inserted into the pre-trained image diffusion models.
VideoBooth~\cite{jiang2024videobooth} further employs cross-frame attention to preserve the fine-grained visual appearance of the customized subject.
Furthermore, CustomVideo~\cite{wang2024customvideo} fine-tunes cross-attention layers on all involved subjects simultaneously to customize multiple subjects within a scene.
However, these approaches merely focus on the customization of the \textit{static subject}. They are limited to offering users or video creators the ability to personalize their desired \textit{dynamic motions} (e.g., specific dancing styles) into output videos, severely hampering the flexibility of video content customization.

On the other hand, to empower the users with the controllability of \emph{dynamic motion}, recent methods~\cite{jeong2024vmc, zhao2023motiondirector, ren2024customizeavideo, wu2025motionmatcher, materzynska2024newmove} have designed modules to capture motion patterns from the conditioned reference videos.
For instance, Customize-A-Video~\cite{ren2024customizeavideo} fine-tunes low-rank adaptation (LoRA)~\cite{hu2021lora} inserted into temporal attention layers to learn the desired motion pattern. Similarly, MotionDirector~\cite{zhao2023motiondirector} employs LoRA to learn motion by using an objective that captures the differences between an anchor frame and the other frames.
However, simply tuning temporal modules without properly disentangling motion information from reference videos causes severe \emph{appearance leakage} issues, resulting in the derived motion patterns that cannot be applied with arbitrary subject identities. Moreover, without guidance for subjects and motion \emph{composition}, the model struggles to precisely control the interaction among these customized video concepts. As a result, the aforementioned methods can only handle \emph{single} (i.e., subject or motion) concept customization. Jointly customizing \emph{multiple} video concepts by free-form prompts that describe multiple subjects \emph{and} desired motion patterns remains a challenging and unsolved problem.

In this paper, we propose \ours{}, a unified framework for video content customization that enables controllability over subject identities and motion patterns. \ours{} involves subject and motion LoRAs to capture respective information from user-provided images and videos. To ensure the motion LoRAs would not be contaminated by visual appearance, we introduce an appearance-agnostic motion learning approach, which isolates motion patterns from reference videos. More specifically, we employ negative classifier-free guidance~\cite{gandikota2023esd, huang2023receler} conditioned on the visual appearance, effectively disentangling motion from appearance details. With the learned subject and motion LoRAs, we introduce a spatial-temporal collaborative composition scheme to guide interactions among multiple subjects in the desired motion pattern. We advance gradient-based fusion and spatial attention regularization to absorb the multi-subject information while encouraging distinct spatial arrangements of subjects. By iteratively guiding the generation process using subject and motion LoRAs, \ours{} synthesizes output videos with enhanced user control and spatiotemporal coherence.        

We now summarize the contributions of this work below:
\begin{itemize}
  \item We propose \textbf{\ours{}}, a unified framework that first enables video concept customization for multiple subject identities and their interactive motion.
  \item We introduce a novel appearance-agnostic motion learning by advancing negative classifier-free guidance to disentangle underlying motion patterns from appearance.
  \item We develop a spatial-temporal collaborative composition scheme to compose the obtained multi-subject and motion LoRAs for generating coherent multi-subject interactions in the desired motion pattern.
\end{itemize}
\begin{figure*}[t!]
  \centering
  \includegraphics[width=0.85\textwidth]{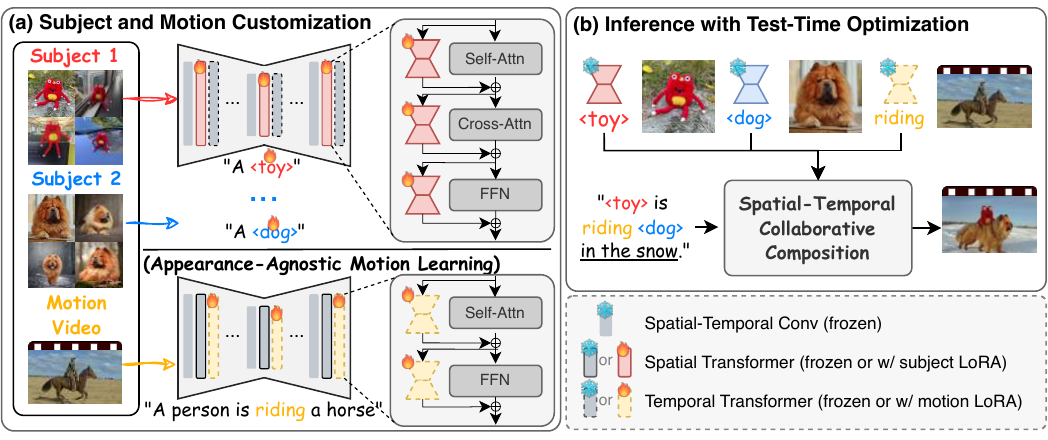}

  \caption{
    \textbf{Overview of \ours{}.}
    (a) Given images of multiple subjects and a reference video with desirable motion, \ours{} advances LoRAs to capture the knowledge of \textit{visual appearances} and \textit{appearance-agnostic motion} information, respectively. (b) With a text prompt relating the aforementioned visual and motion concepts, our \emph{spatial-temporal collaborative composition} refines the input noisy latent $x_t$ for generating videos matching the desirable visual and motion information.
  }

  \label{fig:overview}

\end{figure*}

\section{Related Works}
\label{sec:related_works}

\subsection{Text-to-Video Generation} Recently, text-to-video generation has made remarkable progress. It has evolved from early approaches based on Generative Adversarial Networks (GANs)~\cite{tulyakov2018mocogan, vondrick2016videogan, saito2017tgan, skorokhodov2022styleganv} and autoregressive transformers~\cite{wu2021godiva, yan2021videogpt, weissenborn2019scalingvideo, hong2022cogvideo} to recent diffusion models~\cite{ho2022vdm, ho2022imagenvideo, singer2022makeavideo, blattmann2023align, chen2023videocrafter1, wang2023modelscope}, which have significantly improved both the quality and diversity of generated videos. Pioneering works such as VDM~\cite{ho2022vdm} and ImagenVideo~\cite{ho2022imagenvideo} modeled video diffusion processes in pixel space, while LVDM~\cite{he2022lvdm} and VideoLDM~\cite{blattmann2023align} modeled in latent space to optimize computational efficiency. To overcome the challenge of lacking paired video-text data, Make-A-Video~\cite{singer2022makeavideo} utilizes a text-to-image prior to achieve text-to-video generation. On the other hand, open-source models like VideoCrafter~\cite{chen2023videocrafter1}, ModelScopeT2V~\cite{wang2023modelscope}, and ZeroScope~\cite{zeroscope} incorporate spatiotemporal blocks to enhance text-to-video generation, demonstrating notable capabilities for producing high-fidelity videos. These powerful text-to-video diffusion models have driven advancements in customized content generation.

\subsection{Video Content Customization}
\paragraph{Subject Customization.} In recent years, customized generation has gained considerable attention, particularly in image synthesis~\cite{gal2022textualinversion, ruiz2023dreambooth, kumari2023customdiffusion}. Building on these advances, recent efforts have increasingly focused on video subject customization~\cite{wang2024customvideo, jiang2024videobooth, guo2023animatediff, wei2024dreamvideo, chen2024disenstudio, chen2023videodreamer}, which is more challenging due to the need for generating subjects in dynamic scenes.
For example, AnimateDiff~\cite{guo2023animatediff} inserts additional motion modules into pre-trained image diffusion models, enabling the animation of custom subjects.
Furthermore, VideoBooth~\cite{jiang2024videobooth} employs cross-frame attention mechanisms to preserve the fine-grained visual appearance of the customized subject.
Recently, CustomVideo~\cite{wang2024customvideo} fine-tunes the cross-attention layers on all involved subjects to achieve multi-subject customization.
However, these methods tend to produce slight subject movements~\cite{wei2024dreamvideo, wu2024motionbooth}, lacking the controllability by users to enable precise control over motion.

\paragraph{Motion Customization.} Given a few reference videos describing a target motion pattern, motion customization~\cite{jeong2024vmc, zhao2023motiondirector, ren2024customizeavideo, wei2024dreamvideo, wu2025motionmatcher, materzynska2024newmove} aims to generate videos that replicate the target motion.
For instance, Customize-A-Video~\cite{ren2024customizeavideo} fine-tunes low-rank adaptation (LoRA)~\cite{hu2021lora} integrated into temporal attention layers to capture specific motion patterns from reference videos. Similarly, MotionDirector~\cite{zhao2023motiondirector} learns motion by fine-tuning LoRA to captures the differences between an anchor frame and the other frames, effectively transferring dynamic behaviors into the generated video content. 
Very recently, DreamVideo~\cite{wei2024dreamvideo} explores the customization of a single subject performing specific motions by employing ID and motion adapters, which are separately appended to the spatial and temporal layers.
However, the appearance leakage issue and the lack of proper guidance for subject and motion composition hamper these methods from generating videos with multiple subjects interacting. Thus, the flexibility of customizing video content with arbitrary subjects and motion patterns is strictly limited.
To empower users with enhanced controllability over video concepts of subject and motion, we employ a unique \ours{} framework to enable desired interactions among multiple customized subject identities. 

\section{Method}
\label{sec:method}

\paragraph{Problem Formulation.} We first define the setting and notations. Given $N$ subjects, each represented by 3-5 images denoted as $x_{s,i}$ for the $i$-th subject (omitting the individual image index for simplicity), a reference interactive motion video $x_m$, and a user-provided text prompt $c_{tgt}$, our goal is to generate a video based on $c_{tgt}$ in which these $N$ subjects interact according to the motion pattern.

To tackle the above problem, we propose \ours{}, a unified framework for customizing multiple subjects and interactive motions for text-to-video generation. With a quick review of video diffusion models~(\cref{ssec:preliminary}), we detail how we utilize LoRA modules to learn visual and motion information from input images and reference videos, respectively (\cref{ssec:customization}). Instead of naive combination, a unique spatial-temporal collaborative composition scheme is presented to integrate the learned subjects/motion LoRAs for video generation (\cref{ssec:composition}).

\subsection{Preliminary: Video Diffusion Models}
\label{ssec:preliminary}
Video Diffusion Models (VDMs)~\cite{ho2022vdm, ho2022imagenvideo, singer2022makeavideo, blattmann2023align, he2022lvdm} are designed to generate video by gradually denoising a sequence of noises sampled from a Gaussian distribution~\cite{ho2020ddpm}. Specifically, the diffusion model $\epsilon_\theta$ learns to predict the noise $\epsilon$ added at each timestep $t$, conditioned on the input $c$, which is a text prompt in text-to-video generation. The training objective is simplified to a reconstruction loss:
\begin{equation}\label{eq:diffusion_loss}
    \mathcal{L} = \mathbb{E}_{x,\epsilon,t} \left[ {\lVert \epsilon_\theta(x_t, c, t) - \epsilon \rVert}^2_2 \right],
\end{equation}
where noise $\epsilon \in \mathbb{R}^{F \times H \times W \times 3}$ is sampled from $\mathcal{N}(\mathbf{0}, \mathbf{I})$, timestep $t \in \mathcal{U}(0, 1)$, and $x_t = \sqrt{\bar{\alpha}_t} x + \sqrt{1 - \bar{\alpha}_t} \epsilon$ is the noisy input at $t$, with $\bar{\alpha}_t$ being a hyperparameter for controlling the diffusion process~\cite{ho2020ddpm}.
To reduce computational cost, most VDMs~\cite{wang2023modelscope, chen2023videocrafter1, he2022lvdm} encode the input video data $x \in \mathbb{R}^{F \times H \times W \times 3}$ into a latent representation (e.g., derived by a VAE~\cite{kingma2013vae}). For simplicity, we continue to use video data $x$ as the model's input throughout the paper. 

\subsection{Subject and Motion Customization}
\label{ssec:customization}

\begin{figure}[t!]
  \centering
  \includegraphics[width=0.96\columnwidth]{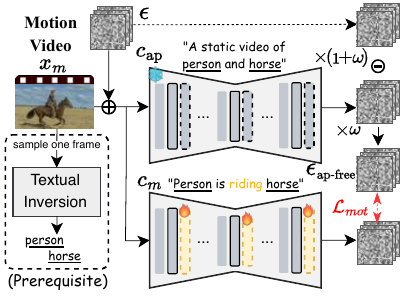}

  \caption{\textbf{Appearance-agnostic motion learning.} By utilizing text prompt emphasizing the appearance information (i.e., $c_{\text{ap}}$), we aim to extract appearance-agnostic motion information via the proposed \textit{negative} classifier-free guidance.}

  \label{fig:motion}

\end{figure}

\begin{figure*}[t!]
  \centering
  \includegraphics[width=1.0\textwidth]{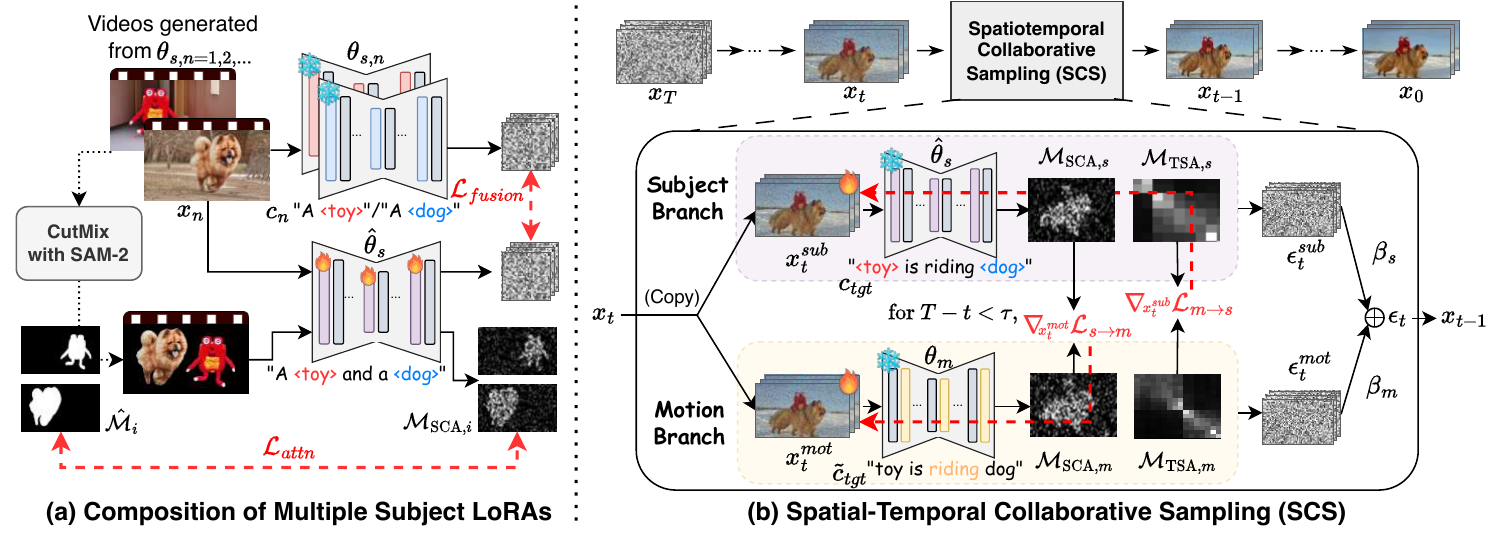}

  \caption{
      \textbf{Spatial-temporal collaborative composition for T2V test-time optimization.} (a) Test-time fusion of subject LoRAs $\hat{\theta}_s$, which employs attention regularization $\mathcal{L}_{attn}$ to ensure appearance preservation of each visual subject.
      (b) Spatiotemporal Collaborative Sampling (SCS) integrates the fused subject LoRA $\hat{\theta}_s$ and the motion LoRA $\theta_m$ by cross-modal alignment, ensuring visual and temporal coherence.
  }

  \label{fig:composition}

\end{figure*}

\paragraph{Learning of Visual Subjects.}

As illustrated at the top of \cref{fig:overview}(a), to capture subject appearance for video generation, we learn a special token (e.g., ``<toy>'') and use a subject LoRA ($\Delta\theta_{s}$) to fine-tune the pre-trained video diffusion model. To avoid interfering with temporal dynamics, the subject LoRA is applied only to the spatial layers of the UNet. The objective is defined as:
\begin{equation}
\label{eq:subject_loss}
    \mathcal{L}_{sub} = \mathbb{E}_{x_s,\epsilon,t} \left[ {\lVert \epsilon_{\theta_{s}}(x_{s,t}, c_s, t) - \epsilon \rVert}^2_2 \right],
\end{equation}
where $x_s \in \mathbb{R}^{1 \times H \times W \times 3}$ is the subject image, $\theta_{s} = \theta + \Delta\theta_{s}$ denotes the parameters of the pre-trained model with the subject LoRA applied, and $c_s$ is the prompt containing the special token (e.g., ``A <toy>'').

However, fine-tuning with image data alone might result in video diffusion models losing their capability in producing motion information. Following~\cite{wu2024motionbooth}, we leverage an auxiliary video dataset $\mathcal{D}_{aux}$ (e.g., Panda70M~\cite{chen2024panda}) to regularize fine-tuning while preserving the pre-trained motion prior.
More precisely, given video-caption pair ($x_{aux}$, $c_{aux}$) sampled from $\mathcal{D}_{aux}$, the regularization loss is defined as:
\begin{equation}
\label{eq:regularization_loss}
    \mathcal{L}_{reg} = \mathbb{E}_{x_{aux},\epsilon,t} \left[ {\lVert \epsilon_{\theta_{s}}(x_{aux,t}, c_{aux}, t) - \epsilon \rVert}^2_2 \right].
\end{equation}
Thus, the overall objective is then defined as:
\begin{equation}
\label{eq:subject_total}
    \mathcal{L} = \mathcal{L}_{sub} + \lambda_1 \mathcal{L}_{reg},
\end{equation}
where $\lambda_1$ is the hyperparameter that control the weight of regularization loss. Optimizing this objective captures the subject appearance while preserving the motion prior.
With our training objective, we are able to allow customization for user-provided subject identities without compromising VDM's capability. However, the tuned VDM remains challenging in precisely controlling motion patterns from reference videos, restricting user's flexibility and control. 

\paragraph{Learning of Appearance-Agnostic Motion.}
\label{par:motion}

To learn the desired motion pattern from the reference video $x_m$, a naive strategy is to fine-tune a motion LoRA and inject it into the UNet’s temporal layers (i.e., $\Delta\theta_{m}$ at the bottom of \cref{fig:overview}(a)). However, direct applying the standard diffusion loss in \cref{eq:diffusion_loss} would result in a \emph{appearance leakage} issue, wherein the motion LoRA inadvertently captures the appearance of subjects from the reference video. This entanglement of subject appearance and motion hinders the ability to apply the learned motion patterns to new subjects.

To address this problem, we propose a novel \emph{appearance-agnostic} objective, as shown in \cref{fig:motion}, which effectively isolates motion patterns from the reference video. 
Inspired by concept erasing methods of~\cite{gandikota2023esd, huang2023receler}, we advance \textit{negative classifier-free guidance} conditioned on the visual subject appearances, focusing on eliminating appearance information during motion learning. This would ensure that the motion LoRA focuses exclusively on motion dynamics.

To achieve this, we first learn special tokens for the subjects in the reference video (e.g., ``person'' and ``horse'' in \cref{fig:motion}) by applying textual inversion~\cite{gal2022textualinversion} on a single frame sampled from the reference video. This captures subject appearance while minimizing motion influence, effectively decoupling appearance from motion. With the above special tokens, we train a motion LoRA using an appearance-agnostic objective that employs negative guidance to suppress appearance information, enabling the motion LoRA to learn motion patterns independently of subject appearances. More specifically, the training objective is defined as:
\begin{multline}
\label{eq:motion_loss}
    \mathcal{L}_{mot} = \mathbb{E}_{x_m,\epsilon,t} \left[ {\lVert \epsilon_{\theta_{m}}(x_{m,t}, c_m, t) - \epsilon_{\text{ap-free}} \rVert}^2_2 \right] \\
    \text{where}\ \epsilon_{\text{ap-free}} =\ (1+\omega) \epsilon - \omega \epsilon_{\theta}(x_{m,t}, c_{\text{ap}}, t).
\end{multline}
Note that $\epsilon_{\text{ap-free}}$ is the negatively guided \emph{appearance-free} noise, $\omega$ is the hyperparameter controlling the guidance strength, and $c_{m}$ and $c_{\text{ap}}$ describe the motion and the static subject appearances, respectively (e.g., ``Person riding a horse'' and ``A \textit{static} video of person and horse'').

By optimizing~\cref{eq:motion_loss}, motion LoRA learns motion patterns independent of subject appearances. This disentanglement is crucial for composing multiple subjects with customized motions, as we will discuss later.

\subsection{Spatial-Temporal Collaborative Composition}
\label{ssec:composition}

With multiple subject LoRAs and an interactive motion LoRA obtained, our goal is to generate videos where these subjects interact using the desired motion pattern. However, combining LoRAs with distinct properties (i.e., visual appearance vs. spatial-temporal motion) is not a trivial task. In our work, we propose a test-time optimization scheme of \emph{spatial-temporal collaborative composition}, which enables collaboration between the aforementioned LoRAs to generate videos with the desired appearance and motion properties. We now discuss the proposed scheme below.

\paragraph{Composition of Multiple Subject LoRAs.}
\label{par:fusion}
We first discuss how we perform fusion of LoRAs describing different visual subject information. We employ gradient-based fusion~\cite{gu2024mixofshow} to distill the distinct identities from each subject LoRA into a single fused LoRA. That is, given multiple LoRAs, denoted as $\Delta\theta_{s,1}, \Delta\theta_{s,2}, \ldots, \Delta\theta_{s,N}$, where $N$ is the number of subjects and each LoRA corresponds to a specific subject, our goal is to learn a fused LoRA $\Delta\hat{\theta}_{s}$ that is able to generate video featuring multiple subjects.

To achieve this, we aim to enforce the fused LoRA $\Delta\hat{\theta}_{s}$ to generate consistent videos with each specific subject LoRA. To be more precise, we optimize $\Delta\hat{\theta}_{s}$ by matching the predicted noise between the fused LoRA and the subject-specific one. The multi-subject fusion objective $\mathcal{L}_{fusion}$ is formulated as follows,  
\begin{multline}
\label{eq:fusion_loss}
    \mathcal{L}_{fusion} = \frac{1}{N} \sum\nolimits_{n=1}^{N} \mathbb{E}_{x_n,\epsilon,t} \left[ {\lVert \epsilon_{\hat{\theta}_{s}}(x_{n,t}, c_n, t) - \epsilon_n \rVert}^2_2 \right],\\
    \text{where}\ \epsilon_n = \epsilon_{\theta_{s,n}}(x_{n,t}, c_n, t).
\end{multline}
Here, $x_n$ is the video generated by $\theta_{s,n}$, and $c_n$ is the corresponding prompt for the $n$-th subject.

Moreover, to encourage different subject identities to be properly arranged, we further introduce \textit{spatial attention regularization} $\mathcal{L}_{attn}$ to explicitly guide the model's attention to focus on the correct subject regions.
Specifically, as illustrated in \cref{fig:composition}(a), we randomly sample and segment two subjects by Grounded-SAM2~\cite{ren2024groundedsam,ravi2024sam2}, and then combine the segmented subjects into a CutMix-style~\cite{yun2019cutmix,han2023svdiff} video. We then formally define $\mathcal{L}_{attn}$ as:
\begin{equation}
\label{eq:attn_reg}
    \mathcal{L}_{attn} = \frac{1}{2} \sum_{i=1}^{2} {\lVert \mathcal{M}_{\text{SCA},i} - \hat{\mathcal{M}}_i \rVert}^2_2,
\end{equation}
where $\mathcal{M}_{\text{SCA},i}$ is the spatial cross-attention map of the $i$-th sampled subject, and $\hat{\mathcal{M}}_i$ is the corresponding ground-truth segmentation mask.
Therefore, the overall objective for deriving the multi-subject LoRA is defined as:
\begin{equation}
\label{eq:fusion_total}
    \mathcal{L} = \mathcal{L}_{fusion} + \lambda_2 \mathcal{L}_{attn},
\end{equation}
where $\lambda_2$ controls the weight of the attention loss. Note that, we only require merging multiple subjects at once. Once the fused LoRA $\hat{\theta}_s$ is obtained, we are able to generate videos with arbitrary motion patterns, as detailed below. 

\paragraph{Spatial-Temporal Collaborative Sampling (\sampling{})}

\begin{figure*}[t]
    \centering
    \includegraphics[width=1.0\textwidth]{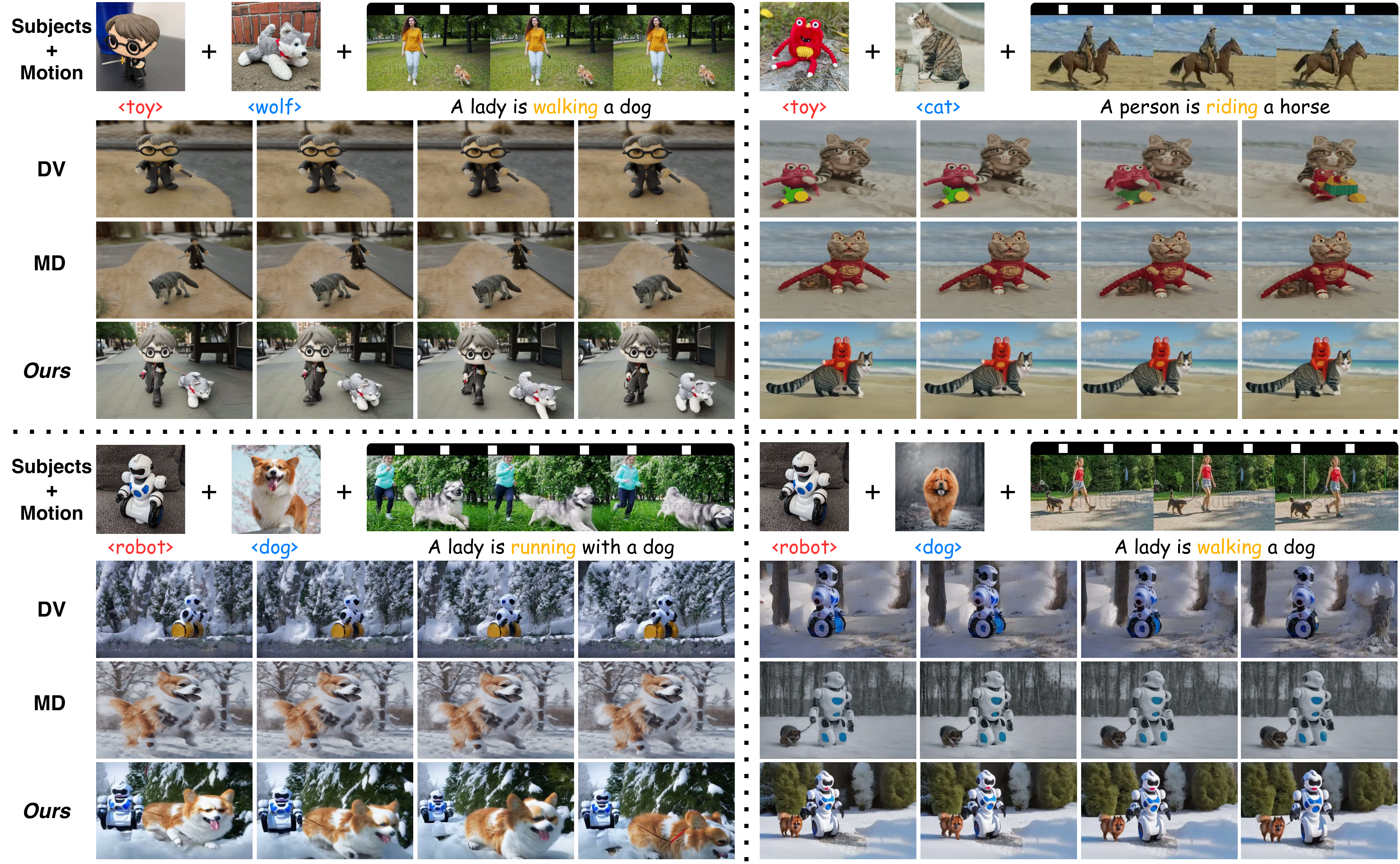}

    \caption{\textbf{Qualitative comparisons of different customization methods.} The subject images and the reference motion video are listed at the top of the figure. DV and MD refer to DreamVideo~\cite{wei2024dreamvideo} and MotionDirector~\cite{zhao2023motiondirector}, respectively. Please refer to the supplementary materials for the complete input prompts used for customization (e.g., describing the background, etc.).}
    \label{fig:qualitative_result}

\end{figure*}

To further integrate the motion-based LoRA, $\Delta\theta_m$, with the aforementioned fused visual subject LoRA, $\Delta\hat{\theta}_s$, we propose a novel \emph{spatial-temporal collaborative sampling} (\sampling{}) technique to effectively control and guide interactions among customized subjects. 
In SCS, we independently sample and integrate noise from the subject branch and the motion branch. To encourage alignment during early timesteps, we introduce a collaborative guidance mechanism where spatial and temporal attention maps from both branches mutually refine each other's input latents. This mutual alignment enables both branches to align effectively, leading to more coherent integration of customized subjects and their interaction.

As illustrated in \cref{fig:composition}(b), given a noised video input $x_t$, we duplicate it into $x_{t}^{sub}$ and $x_{t}^{mot}$ for the subject and motion branches. With $\hat{\theta}_s$ and $\theta_m$ denoting the models with the fused subject LoRA and the motion LoRA applied, respectively, we generate \textit{subject} noise ($\epsilon^{sub}_t$) and \textit{motion} noise ($\epsilon^{mot}_t$) as follows:
\begin{equation}
\begin{split}
    \epsilon^{sub}_t &= \epsilon_{\hat{\theta}_{s}}(x_{t}^{sub}, c_{tgt}, t), \\\ \epsilon^{mot}_t &= \epsilon_{\theta_{m}}(x_{t}^{mot}, \tilde{c}_{tgt}, t),
\end{split}
\end{equation}
where $c_{tgt}$ is the input prompt containing subject special tokens (e.g., ``A <toy> is riding a <dog>''), and $\tilde{c}_{tgt}$ is constructed by replacing the special tokens with their respective superclasses (e.g., ``A toy is riding a dog'').

However, since the subject branch (with only subject LoRA) generates incorrect motion, and the motion branch (with only motion LoRA) produces inaccurate spatial arrangements, directly combining $\epsilon^{sub}_t$ and $\epsilon^{mot}_t$ might result in incomplete information from either modality.
Therefore, we encourage alignment between $\hat{\theta}s$ and $\theta_m$ to produce coherent noise outputs. To achieve this alignment, as depicted in \cref{fig:composition}(b), we consider spatial cross-attention maps ($\mathcal{M}_{\text{SCA}}$), capturing the spatial arrangement of subjects, and temporal self-attention maps ($\mathcal{M}_{\text{TSA}}$), capturing motion dynamics, as demonstrated in previous works~\cite{gu2024mixofshow, ling2024motionclone, wang2024customvideo}.

Specifically, we enforce motion correctness by aligning the temporal self-attention maps of the subject branch with those of the motion branch. Similarly, we ensure accurate spatial arrangements by aligning the spatial cross-attention maps of the motion branch with those of the subject branch. Losses for collaborative guidance are calculated as:
\begin{equation}
\begin{split}
\label{eq:collaborative_guidance}
    \mathcal{L}_{s \rightarrow m} &= {\lVert \mathcal{M}_{\text{SCA},s} - \mathcal{M}_{\text{SCA},m} \rVert}^2_2,\\
    \mathcal{L}_{m \rightarrow s} &= {\lVert \mathcal{M}_{\text{TSA},s} - \mathcal{M}_{\text{TSA},m} \rVert}^2_2,
\end{split}
\end{equation}
where the subscripts $s$ and $m$ indicate the maps are from subject and motion branches, respectively.
Similar to \cite{agarwal2023astar, chefer2023attend_and_excite}, we update $x_{t}^{sub}$ and $x_{t}^{mot}$ as follows:
\begin{equation}
\begin{split}
\label{eq:collaborative_refine}
    x_{t}^{sub} &:= x_{t}^{sub} - \alpha_{t} \nabla_{x_{t}^{sub}} \mathcal{L}_{m \rightarrow s},\\
    x_{t}^{mot} &:= x_{t}^{mot} - \alpha_{t} \nabla_{x_{t}^{mot}} \mathcal{L}_{s \rightarrow m},
\end{split}
\end{equation}
where $\alpha_{t}$ is the step size of the gradient update.
This guidance is applied for the first $\tau$ denoising steps, where $\tau$ is a hyperparameter. Finally, the predicted noise is calculated by $\epsilon_{t} = \beta_{s} \epsilon^{sub}_t + \beta_{m} \epsilon^{mot}_t$ where we set $\beta_{s} = \beta_{m} = 0.5$ for simplicity. We leave more details in Algorithm 1 of our supplementary material.

\section{Experiment}
\label{sec:experiments}

\subsection{Experimental Setup}
\paragraph{Dataset.}\label{par:dataset} To evaluate video customization methods for multi-subject and motion tasks, we collect 6 motion videos from WebVid~\cite{Bain21}, featuring various interactions between people and animals. For each motion, we provide 3 subject pairs from~\cite{ruiz2023dreambooth, kumari2023customdiffusion}, including diverse species such as animals, robots, toys, and plushies, with 4 different background prompts per setting.

\paragraph{Evaluation Metrics.} Following prior works~\cite{wei2024dreamvideo, zhao2023motiondirector, wang2024customvideo}, we evaluate performance using: 1) \textit{CLIP-T}, which measures cosine similarity between generated frames and text prompts using CLIP~\cite{radford2021learning}; 2) \textit{CLIP-I}, which assesses subject identity by comparing CLIP image embeddings of generated frames and target images; 3) \textit{DINO-I}, similar to \textit{CLIP-I}, but using embeddings from DINO~\cite{caron2021emerging}; 4) \textit{Temporal Consistency}~\cite{esser2023structure}, which measures frame-wise consistency by calculating similarity between consecutive frames using CLIP. Additionally, we conduct human evaluations for qualitative assessment.

\paragraph{Comparisons.} We compare our \ours{} with state-of-the-art video customization methods, including DreamVideo~\cite{wei2024dreamvideo} and MotionDirector~\cite{zhao2023motiondirector}, which customize a single subject with motion by applying adapters and LoRAs, respectively. For fair comparisons, we first average the outputs from multiple subject modules and combine them with motion modules for multi-subject and motion customization.

\paragraph{Implementation Details.} For \ours{}, both subject and motion LoRAs are trained for 300 iterations with a rank of 4. We set the learning rates as $10^{-4}$ for LoRAs and $10^{-3}$ for textual embeddings. Hyperparameters $\lambda_1$, $\lambda_2$, and $\omega$ are set to $0.25$, $0.6$, and $0.5$, respectively. For SCS, $\tau = 15$, and $\alpha_t$ starts at $10^4$ and decays by half by the end of denoising, following~\cite{chefer2023attend_and_excite}. For all experiments, we adopt ZeroScope~\cite{zeroscope} as the video diffusion model. Following~\cite{wei2024dreamvideo}, we use a $50$-step DDIM~\cite{song2020ddim} with a guidance scale of $9.0$ to generate $24$-frame videos at $8$ fps, with a resolution of $320\times576$. Please refer to the supplementary material for more details.

\begin{table}[t!]
    \centering
    \resizebox{0.98\columnwidth}{!}{
    \begin{tabular}{lcccc} 
        \toprule
        \textbf{Method} & \textbf{CLIP-T}& \textbf{CLIP-I}& \textbf{DINO-I} & \tabincell{c}{\textbf{T. Cons.} }\\ 
        \midrule
        DreamVideo~\cite{wei2024dreamvideo} & 0.582 & 0.605 & 0.197 & 0.972 \\ 
        MotionDirector~\cite{zhao2023motiondirector} & 0.656 & 0.634 & 0.370 & \textbf{0.987} \\ 
        \midrule
        \textbf{\ours{} (ours)} & \textbf{0.662} & \textbf{0.670} & \textbf{0.407} & 0.983 \\ 
        \bottomrule
    \end{tabular}
    }
    \caption{\textbf{Quantitative comparison on multi-subject and motion customization.} We follow \cite{wei2024dreamvideo, zhao2023motiondirector} to adopt metrics including CLIP-Text Alignment (\textit{CLIP-T}), CLIP-Image Alignment (\textit{CLIP-I}), DINO-Image Alignment (\textit{DINO-I}), and Temporal Consistency (T. Cons.).}
    \label{tab:quantitative_compare}
\end{table}

\begin{figure}[t!]
  \centering
  \includegraphics[width=1.0\linewidth]{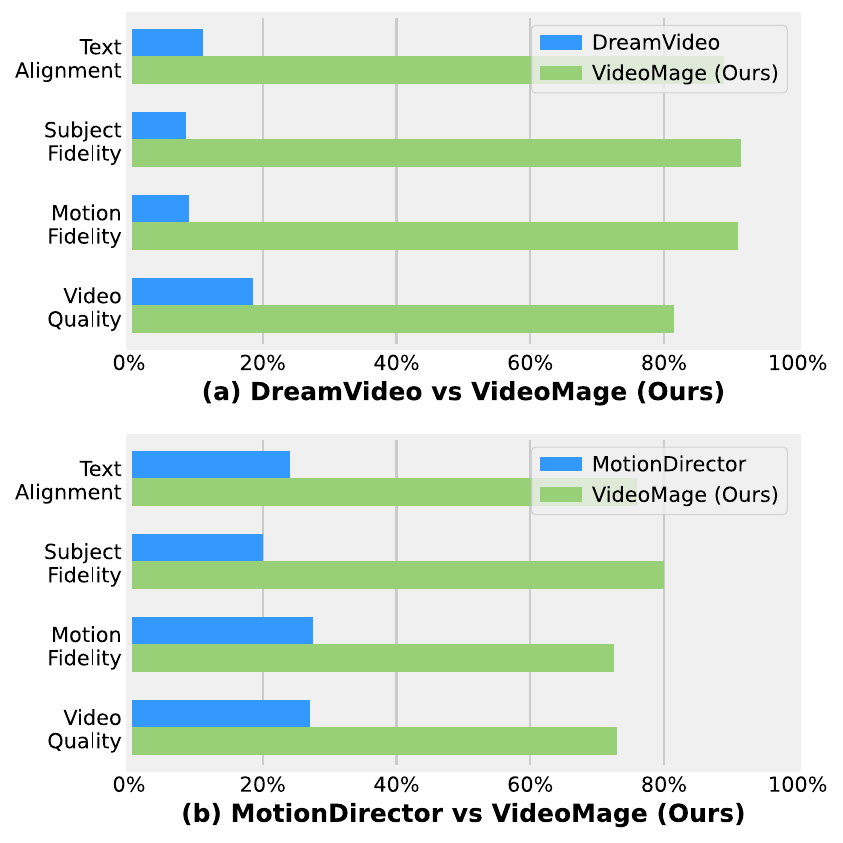}
  \caption{\textbf{Human preference study.} Our \ours{} consistently achieves the best human preference compared to DreamVideo~\cite{wei2024dreamvideo} and MotionDirector~\cite{wu2024motionbooth}.}
  \label{fig:user_study}
\end{figure}

\subsection{Main Results}

\paragraph{Qualitative Results.}
In \cref{fig:qualitative_result}, we illustrate examples of customized video generation that combine various user-provided subject images with a specific motion reference video. As we can observe, both DreamVideo and MotionDirector suffer from significant \emph{appearance leakage} and \emph{attribute mixing} issues, struggling to correctly arrange multiple subjects to follow the referenced motion pattern.
For instance, in the lower right corner, the black dog's appearance from the motion video is unintentionally transferred into MotionDirector's output, while in the DreamVideo's output in the lower left corner, the color attribute of the <dog> is incorrectly mixed with <robot>, resulting in undesirable visual details. Moreover, both methods fail to establish the intended interactions among subjects, falling short of capturing the nuanced dynamics between them. In contrast, our \ours{} effectively addresses these challenges, preserving subject identities, preventing appearance leakage, and successfully achieving the desired interactions between subjects in the generated video.

\paragraph{Quantitative Results.}
We conduct quantitative evaluations on our collected multi-subject and motion dataset (as described in \cref{par:dataset}). With a total of 72 combinations of subjects, motions, and backgrounds, we generate 10 videos for each combination and evaluate them using four metrics. As shown in \cref{tab:quantitative_compare}, our \ours{} generates videos that better preserve the subjects' identities, outperforming the state-of-the-art method, MotionDirector, by $5.7\%$ and $10\%$ for \textit{CLIP-I} and \textit{DINO-I}, respectively. Additionally, \ours{} achieves the highest \textit{CLIP-T} performance and is comparable to SOTA for \textit{Temporal Consistency}, demonstrating its ability to generate coherent videos that align closely with the text prompts.

\paragraph{User Study.} To further assess the effectiveness of our method, we conduct a human preference study to evaluate our method against DreamVideo~\cite{wei2024dreamvideo} and MotionDirector~\cite{zhao2023motiondirector}. In this study, participants are given reference subject images and a motion video, along with two customized videos generated by our \ours{} and a comparison method, respectively. Participants are asked to choose their preferred video based on four criteria: \textit{Text Alignment} (how well the video matches the prompt), \textit{Subject Fidelity} (how closely the subjects match the reference images without incorrect attribute mixing), \textit{Motion Fidelity} (how accurately the motion reflects the reference video), and \textit{Video Quality} (smoothness and absence of flicker). A total of 360 videos are generated, with 25 participants involved in the evaluation. As shown in \cref{fig:user_study}, our \ours{} was preferred by participants across all criteria.

\begin{figure}[t!]
  \centering
  \includegraphics[width=1.0\columnwidth]{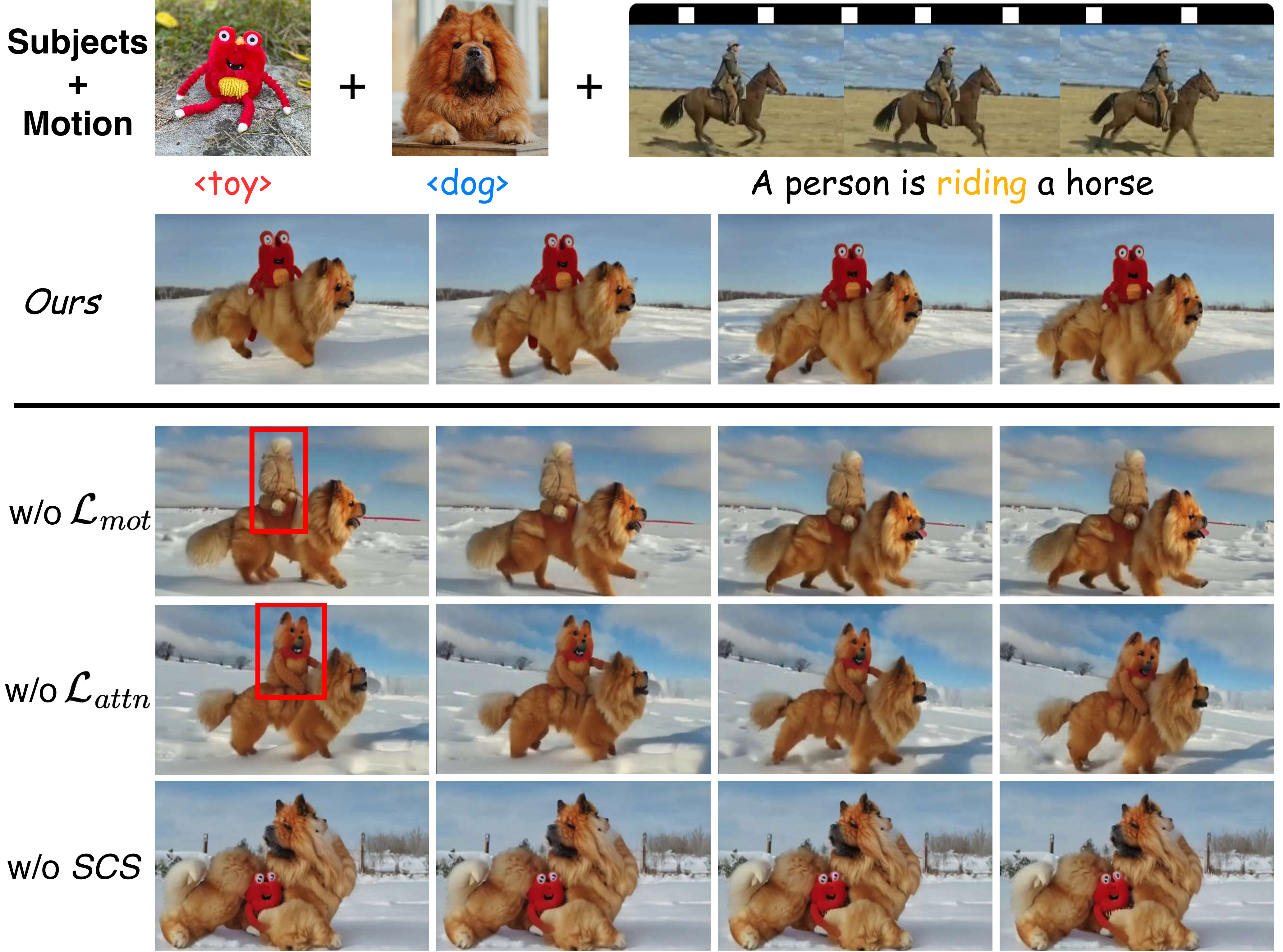}

  \caption{\textbf{Ablation study.} Ablation of different components in \ours{}. Red boxes indicate appearance leakage or attribute binding issues after removing the specified components.}

  \label{fig:ablation_study}

\end{figure}

\begin{table}[t!]
    \centering
    \resizebox{0.93\columnwidth}{!}{
    \begin{tabular}{lcccc} 
        \toprule
        \textbf{Method} & \textbf{CLIP-T}& \textbf{CLIP-I}& \textbf{DINO-I} & \tabincell{c}{\textbf{T. Cons.} }\\ 
        \midrule
        \textbf{\ours{}} & \textbf{0.662} & \textbf{0.670} & \textbf{0.407} & \textbf{0.983} \\
        \midrule
        w/o $\mathcal{L}_{mot}$ & 0.639 & 0.651 & 0.362 & 0.976 \\
        w/o $\mathcal{L}_{attn}$ & 0.626 & 0.647 & 0.358 & 0.978 \\
        w/o SCS & 0.601 & 0.612 & 0.234 & 0.982 \\
        \bottomrule
    \end{tabular}
    }
    \caption{
        \textbf{Quantitative ablation study.} Ablation of our proposed objectives/sampling strategy in \ours{}.
    }
    \label{tab:quantitative_ablation}
\end{table}

\subsection{Ablation Studies}
In \cref{fig:ablation_study}, we present a qualitative ablation study to analyze the contributions of different components in our proposed \ours{}. For the $\text{w/o}\ \mathcal{L}_{mot}$ setting, we learn the motion pattern using the standard diffusion loss (i.e., \cref{eq:diffusion_loss}) instead of the appearance-agnostic objective $\mathcal{L}_{mot}$. As a result, we observe severe \emph{appearance leakage}, where the appearance of the person from the reference motion video is unintentionally transferred to the generated output. In the $\text{w/o}\ \mathcal{L}_{attn}$ setting, we exclude attention regularization during multi-subject fusion, which leads to \emph{attribute binding} issue with mixed attributes between the subjects (e.g., the <toy> unintentionally looks like a combination of <toy> and <dog>). Lastly, in the $\text{w/o SCS}$ setting, we directly combine $\hat{\theta}_s$ and $\theta_m$ in the video diffusion model for inference, which struggles to properly arrange subjects with the desired interactive motion.  Additionally, we further assess the impact of each of our proposed objectives/modules in \cref{tab:quantitative_ablation}. We adopt four metrics to evaluate caption-video similarity (\textit{CLIP-T}), customized subject fidelity (\textit{CLIP-I, DINO-I}), and frame-wise consistency (\textit{T. Cons.}). From the above ablation studies, we successfully verify the effectiveness of our designs.
\section{Conclusion}
\label{sec:conclusion}

In this paper, we proposed a unified framework \ours{} to enable video customization of text-to-video diffusion models among user-provided subject identities and the desired motion patterns. In \ours{}, we employ multi-subject and appearance-agnostic motion learning to derive the customized LoRAs, while presenting a spatial-temporal collaborative composition scheme to mutually align subject and motion components for synthesizing videos with sufficiently visual and temporal fidelity. We conducted extensive quantitative and qualitative evaluations on \ours{}, validating its superior controllability over previous video customization methods.

\paragraph{Acknowledgment} This work is supported in part by the National Science and Technology Council via grant NSTC 113-2634-F-002-005 and NSTC 113-2640-E-002-003, and the Center of Data Intelligence: Technologies, Applications, and Systems, National Taiwan University (grant nos.114L900902, from the Featured Areas Research Center Program within the framework of the Higher Education Sprout Project by the Ministry of Education (MOE) of Taiwan). We also thank the National Center for High-performance Computing (NCHC) for providing computational and storage resources.
{
    \small
    \bibliographystyle{ieeenat_fullname}
    \bibliography{main}
}
\clearpage
\maketitlesupplementary

\section{Limitation and Future Work}
While our method effectively customizes multiple subjects and their motions in videos, it currently lacks the capability to customize long motions and generate corresponding extended videos (e.g., minute-long videos). This limitation is common across all existing methods, as customizing longer videos requires significant computational resources, either during training or inference.

To address this, future work will explore integrating long video generation techniques or training-free customization methods to enable longer customized video generation. By leveraging advancements in efficient video synthesis capable of handling long video sequences, we aim to improve the generation of longer and more intricate customized video content.

\section{Additional Experimental Setup}

\begin{algorithm}[b]
    \caption{\small Spatial-Temporal Collaborative Sampling (SCS)}
    \label{alg:scs_algo}
    \textbf{Model:} Pre-trained video diffusion model $\theta$, fused multi-subject LoRA $\Delta\hat{\theta}_s$, motion LoRA $\Delta\theta_m$\\
    \textbf{Input:} Target text prompt $c_{tgt}$ (w/ subjects' special tokens) and $\tilde{c}_{tgt}$ (w/o special tokens), initial noise map $x_T$\\
    \textbf{Output:} Sampled video $x_0$
    \begin{algorithmic}[1]
        \FOR{$t = T, T-1, ..., 1$}
            \STATE Duplicate $x_t$ to create $x_t^{sub}$ and $x_t^{mot}$;
            \STATE \label{alg:scs_algo:subject_noise} $\epsilon_t^{sub} = \epsilon_{\hat{\theta}_s}(x_t^{sub}, c_{tgt}, t)$; \hfill \COMMENT{Subject branch noise}
            \STATE \label{alg:scs_algo:motion_noise} $\epsilon_t^{mot} = \epsilon_{\theta_m}(x_t^{mot}, \tilde{c}_{tgt}, t)$; \hfill \COMMENT{Motion branch noise}
            \IF{$T - t < \tau$}
                \STATE \textcolor{blue}{\textit{/* Collaborative Guidance */} }
                \STATE $\mathcal{L}_{s \rightarrow m} = \lVert \mathcal{M}_{\text{SCA},s} - \mathcal{M}_{\text{SCA},m} \rVert^2_2$;
                \STATE $\mathcal{L}_{m \rightarrow s} = \lVert \mathcal{M}_{\text{TSA},s} - \mathcal{M}_{\text{TSA},m} \rVert^2_2$;
                \STATE $x_{t}^{sub} := x_{t}^{sub} - \alpha_{t} \nabla_{x_{t}^{sub}} \mathcal{L}_{m \rightarrow s}$;
                \STATE $x_{t}^{mot} := x_{t}^{mot} - \alpha_{t} \nabla_{x_{t}^{mot}} \mathcal{L}_{s \rightarrow m}$;
                \STATE Execute lines \ref{alg:scs_algo:subject_noise} and \ref{alg:scs_algo:motion_noise} to get updated $\epsilon_t^{sub}$ and $\epsilon_t^{mot}$;
            \ENDIF
            \STATE $\epsilon_t = \beta_s \epsilon_t^{sub} + \beta_m \epsilon_t^{mot}$ and obtain $x_{t-1}$;
        \ENDFOR
        \STATE \textbf{Return} $x_0$;
    \end{algorithmic}
\end{algorithm}

\subsection{Additional Implementation Details}
\paragraph{Appearance-Agnostic Motion Learning.} As described in \cref{par:motion}, we employ Textual Inversion~\cite{gal2022textualinversion} to obtain the special tokens representing subject appearances from the reference motion video for our proposed \emph{appearance-agnostic motion learning}. Specifically, we extract a single frame from the reference video and use Grounded-SAM~\cite{ren2024groundedsam} to obtain segmentation masks for each subject. We then crop each subject based on its corresponding mask and learn a special token (i.e., embedding) for each subject using \cref{eq:subject_total}. This approach ensures that the learned tokens accurately reflect the visual identities of the subjects without incorporating any motion information, which is crucial for the appearance-agnostic motion learning phase.

\begin{figure}[t!]
    \centering
    \includegraphics[width=1.0\columnwidth]{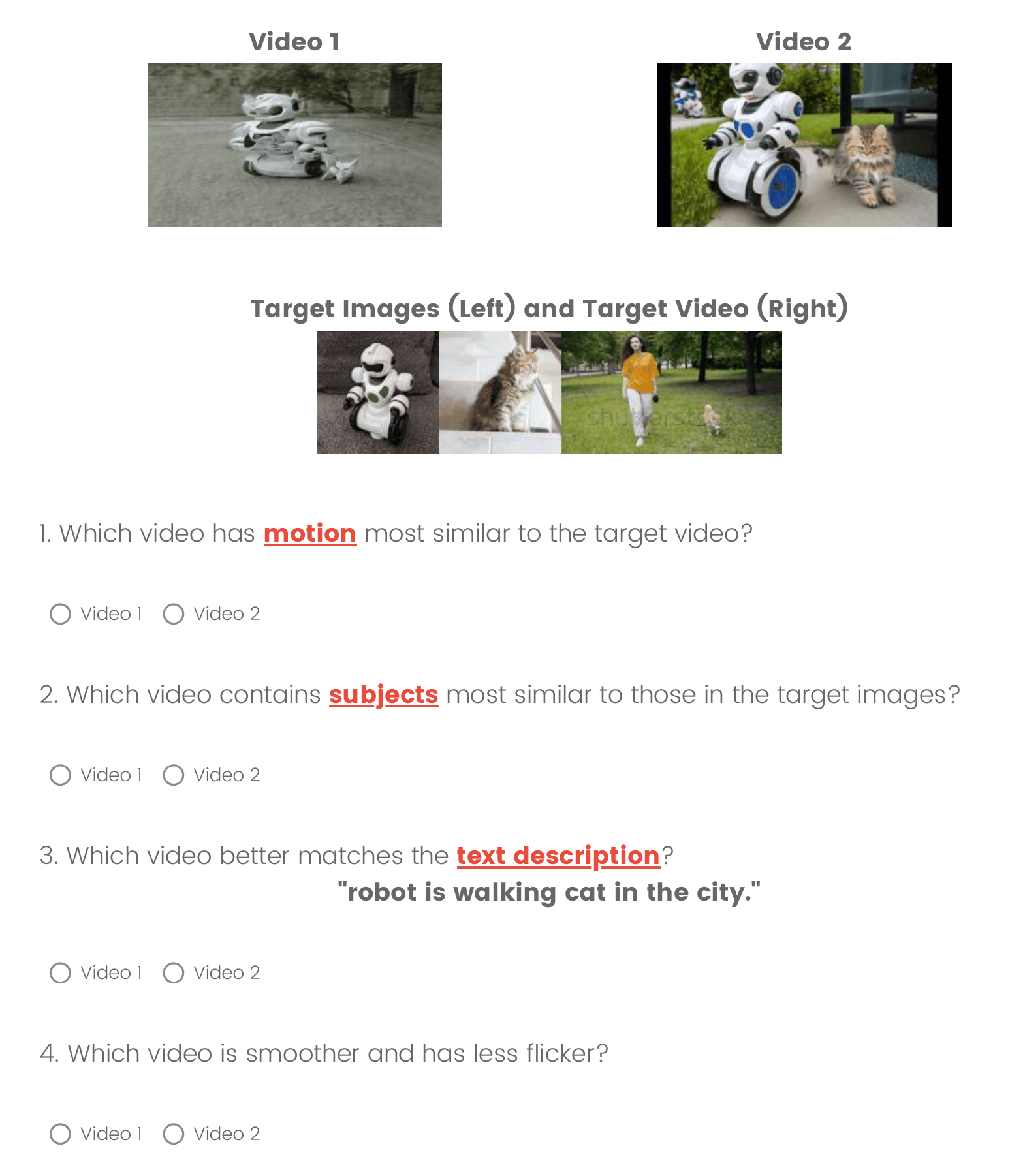}
    \caption{
        \textbf{User study interface.} Given two generated videos, reference subject images, and a reference motion video, participants compare the generated videos based on \textit{Motion Fidelity}, \textit{Subject Fidelity}, \textit{Text Alignment}, and \textit{Video Quality}.
    }
    \label{fig:user_study_interface}
\end{figure}

\begin{table*}[t!]
    \centering
    \begin{subtable}[t]{0.49\textwidth}
        \centering
        \resizebox{0.77\columnwidth}{!}{
            \begin{tabular}{lcccc}
                \toprule
                $\lambda_1$ & \textbf{CLIP-T} & \textbf{CLIP-I} & \textbf{DINO-I} & \textbf{T. Cons.} \\
                \midrule
                0.1 & 0.658 & 0.667 & 0.398 & 0.981 \\
                0.25 & \textbf{0.662} & \textbf{0.670} & \textbf{0.407} & \textbf{0.983} \\
                1.0 & 0.660 & 0.667 & 0.401 & 0.980 \\
                \bottomrule
            \end{tabular}
        }
        \caption{Weight for video preservation loss $\lambda_1$}
    \end{subtable}
    \hfill
    \begin{subtable}[t]{0.49\textwidth}
        \centering
        \resizebox{0.77\columnwidth}{!}{
            \begin{tabular}{lcccc}
                \toprule
                $\lambda_2$ & \textbf{CLIP-T} & \textbf{CLIP-I} & \textbf{DINO-I} & \textbf{T. Cons.} \\
                \midrule
                0.1 & 0.641 & 0.659 & 0.362 & 0.980 \\
                0.6 & \textbf{0.662} & \textbf{0.670} & \textbf{0.407} & 0.983 \\
                1.0 & 0.656 & 0.665 & 0.402 & \textbf{0.984} \\
                \bottomrule
            \end{tabular}
        }
        \caption{Weight for attention regularization loss $\lambda_2$}
    \end{subtable}


    \begin{subtable}[t]{0.49\textwidth}
        \centering
        \resizebox{1.0\columnwidth}{!}{
            \begin{tabular}{lcc}
                \toprule
                \textbf{Template for $c_{\text{ap}}$} & \textbf{CLIP-I} & \textbf{DINO-I} \\
                \midrule
                ``A static video of <sub1> and <sub2>.'' & \textbf{0.670} & \textbf{0.407} \\
                ``A video of <sub1> and <sub2> being still.'' & 0.664 & 0.405 \\
                ``A video of <sub1> and <sub2>.'' & 0.659 & 0.395 \\
                \bottomrule
            \end{tabular}
        }
        \caption{Template for appearance prompt $c_{\text{ap}}$}
    \end{subtable}
    \hfill
    \begin{subtable}[t]{0.49\textwidth}
        \centering
        \resizebox{0.77\columnwidth}{!}{
            \begin{tabular}{lcccc}
                \toprule
                $\omega$ & \textbf{CLIP-T} & \textbf{CLIP-I} & \textbf{DINO-I} & \textbf{T. Cons.} \\
                \midrule
                0.1 & 0.646 & 0.658 & 0.372 & 0.981 \\
                0.5 & \textbf{0.662} & \textbf{0.670} & \textbf{0.407} & \textbf{0.983} \\
                1.0 & 0.657 & 0.667 & 0.403 & 0.980 \\
                \bottomrule
            \end{tabular}
        }
        \caption{Scale factor of negative guidance $\omega$}
    \end{subtable}


    \begin{subtable}[t]{0.49\textwidth}
        \centering
        \resizebox{0.77\columnwidth}{!}{
            \begin{tabular}{lcccc}
                \toprule
                $\alpha_t$ & \textbf{CLIP-T} & \textbf{CLIP-I} & \textbf{DINO-I} & \textbf{T. Cons.} \\
                \midrule
                $10^3$ & 0.657 & 0.665 & 0.401 & \textbf{0.988} \\
                $10^4$ & \textbf{0.662} & \textbf{0.670} & \textbf{0.407} & 0.983 \\
                $10^5$ & 0.634 & 0.658 & 0.379 & 0.976 \\
                \bottomrule
            \end{tabular}
        }
        \caption{Scale factor of collaborative guidance $\alpha_t$}
    \end{subtable}
    \hfill
    \begin{subtable}[t]{0.49\textwidth}
        \centering
        \resizebox{0.77\columnwidth}{!}{
            \begin{tabular}{lcccc}
                \toprule
                $\tau$ & \textbf{CLIP-T} & \textbf{CLIP-I} & \textbf{DINO-I} & \textbf{T. Cons.} \\
                \midrule
                5 & 0.658 & 0.664 & 0.401 & 0.978 \\
                15 & \textbf{0.662} & \textbf{0.670} & \textbf{0.407} & \textbf{0.983} \\
                30 & 0.657 & 0.664 & 0.399 & 0.975 \\
                \bottomrule
            \end{tabular}
        }
        \caption{Steps of collaborative guidance $\tau$}
    \end{subtable}

    \caption{Ablation studies on various hyperparameters, including the weights for video preservation loss ($\lambda_1$) and attention regularization loss ($\lambda_2$), the template for appearance prompt ($c_{\text{ap}}$), the negative guidance scale factor ($\omega$), the collaborative guidance scale ($\alpha_t$) and steps ($\tau$).}
    \label{tab:combined_ablation}
\end{table*}

\paragraph{Spatial-Temporal Collaborative Composition.} As mentioned in \cref{par:fusion}, we sample and preprocess two single-subject training videos by combining them into a CutMix-style~\cite{yun2019cutmix, han2023svdiff} video for regularizing the LoRA fusion. Specifically, for each video, we use Grounded-SAM2~\cite{ren2024groundedsam, ravi2024sam2} to generate segmentation masks for the subjects. We then crop the subjects from the original frames and place them onto a clean background video. To encourage potential interactions between the subjects, we allow some degree of overlap in their placements. We initialize the fused LoRA $\hat{\theta}_s$ with the average of the subject LoRAs. The training steps range from $250$ to $450$, depending on the subject combination.

For \emph{spatial-temporal collaborative sampling} (SCS), we provide the details in \cref{alg:scs_algo}. Following prior works~\cite{wei2024dreamvideo, zhao2023motiondirector}, we initialize the noise map as $x_T = \sqrt{\beta}\epsilon_{m} + \sqrt{1-\beta}\epsilon$, where $\beta = 0.3$, $\epsilon_{m}$ is the DDIM~\cite{song2020ddim} inverted noise of the motion video, and $\epsilon$ is Gaussian noise sampled from $\mathcal{N}(\mathbf{0}, \mathbf{I})$. This initialization is consistently applied to all comparison methods in all experiments.

\subsection{Human User Study}
In \cref{fig:user_study_interface}, we present the interface for our human preference study. In this study, participants are provided with reference subject images, a reference motion video, and two customized videos: one from our \ours{} method and one from a comparison method (i.e., DreamVideo~\cite{wei2024dreamvideo} or MotionDirector~\cite{zhao2023motiondirector}). They are asked to choose their preferred video based on four questions, each evaluating: \textit{Motion Fidelity}, \textit{Subject Fidelity}, \textit{Text Alignment}, and \textit{Video Quality}. A total of $360$ videos were generated for each method, and $25$ participants participated in the study.

\section{Additional Results}
\subsection{Ablation Studies on Hyperparameter Choices}
\begin{table}[t!]
    \centering
    \resizebox{0.98\columnwidth}{!}{
    \begin{tabular}{lcccc} 
        \toprule
        \textbf{Method} & \textbf{CLIP-T}& \textbf{CLIP-I}& \textbf{DINO-I} & \tabincell{c}{\textbf{T. Cons.} }\\ 
        \midrule
        DisenStudio~\cite{chen2024disenstudio} & 0.661 & 0.658 & 0.381 & 0.842 \\
        CustomVideo~\cite{wang2024customvideo} & \textbf{0.676} & 0.679 & 0.402 & 0.849 \\
        \midrule
        \textbf{\ours{} (ours)} & 0.674 & \textbf{0.681} & \textbf{0.403} & \textbf{0.851} \\
        \bottomrule
    \end{tabular}
    }
    \caption{\textbf{Quantitative comparison on multi-subject customization.} Following \cite{chen2024disenstudio, wang2024customvideo}, we evaluate using CLIP-Text Alignment (\textit{CLIP-T}), CLIP-Image Alignment (\textit{CLIP-I}), DINO-Image Alignment (\textit{DINO-I}), and Temporal Consistency (T. Cons.).}
    \label{tab:multi_subject}
\end{table}

\paragraph{Effect of $\lambda_1$ in subject learning.}
As illustrated in \cref{tab:combined_ablation}(a), a video preservation loss weight of $\lambda_1 = 0.25$ achieves the best performance, while both smaller ($0.1$) and larger ($1.0$) values lead to declines. Thus, we set $\lambda_1 = 0.25$ for all experiments.

\paragraph{Effect of $\lambda_2$ in multi-subject fusion.}
As shown in \cref{tab:combined_ablation}(b), the optimal performance is achieved with the attention regularization loss weight set to $\lambda_2 = 0.6$, whereas smaller ($0.1$) or larger ($1.0$) values lead to reduced performance. Thus, we use $\lambda_2 = 0.6$ in our experiments.

\paragraph{Effect of $c_\text{ap}$ and $\omega$ in appearance-agnostic motion learning.}
As shown in \cref{tab:combined_ablation}(c), we experiment with three different templates for the subject appearance prompt $c_\text{ap}$ used in our appearance-agnostic motion learning. The template, ``A static video of <sub1> and <sub2>,'' achieves the best performance and is therefore used in our experiments. Notably, all three templates outperform the second-best result achieved by MotionDirector~\cite{zhao2023motiondirector}, as presented in \cref{tab:quantitative_compare}. Similarly, in \cref{tab:combined_ablation}(d), we present the ablation study on the scale factor of negative guidance $\omega$. We observe that setting $\omega$ to $0.5$ yields the best results; thus, $\omega = 0.5$ is adopted for all experiments.

\begin{figure*}[b]
    \centering
    \includegraphics[width=0.85\textwidth]{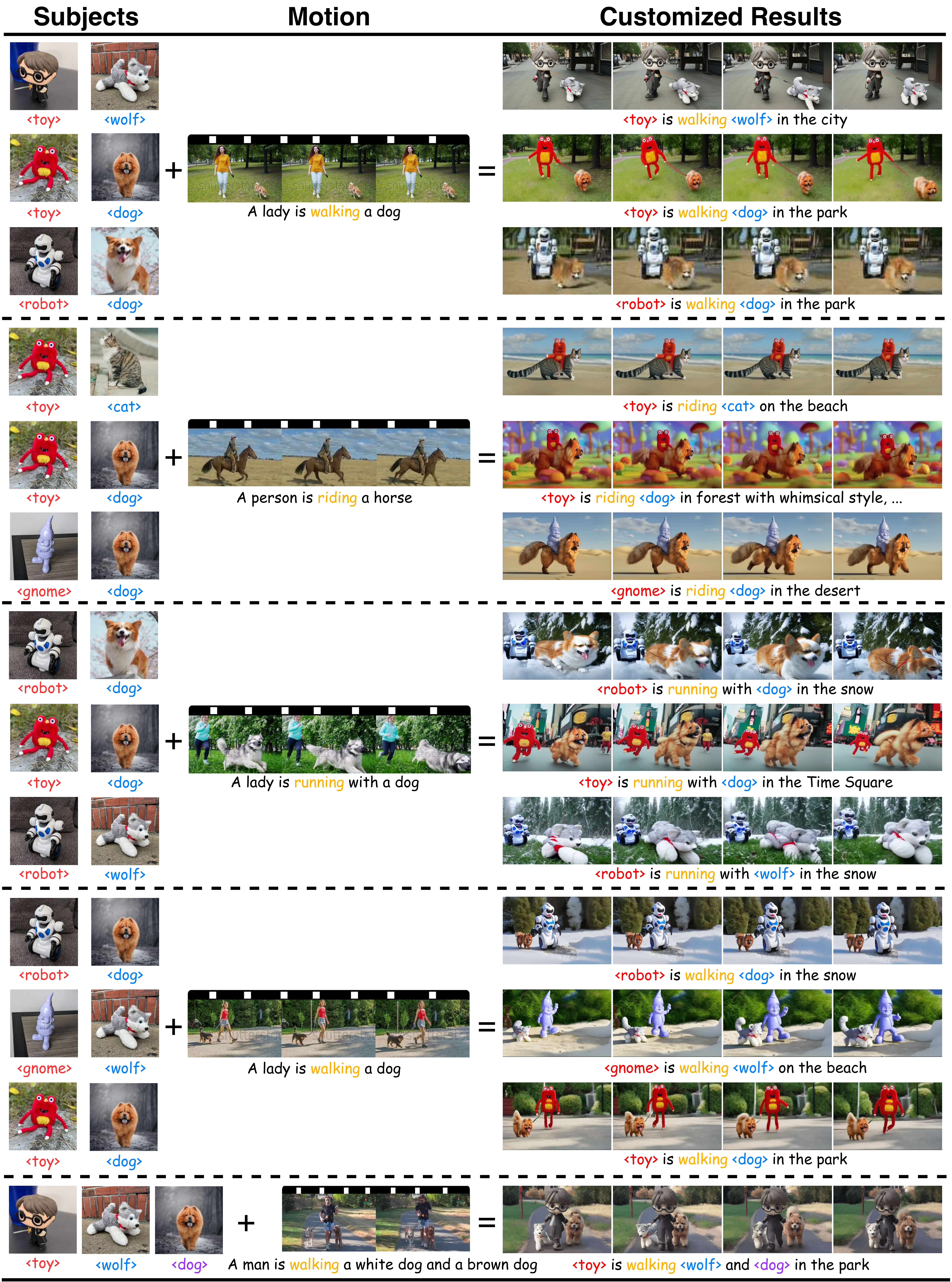}
    
    \caption{\textbf{Additional qualitative results.} Each row presents the subject images, the motion video, the corresponding customized video results, and the input prompt.}

    \label{fig:supp_qualitative}

\end{figure*}

\paragraph{Effect of $\tau$ and $\alpha_t$ in spatial-temporal collaborative sampling.} In \cref{tab:combined_ablation}(e) and \cref{tab:combined_ablation}(f), we ablate the scale factor $\alpha_t$ and steps $\tau$ in our proposed spatial-temporal collaborative sampling, respectively. As shown in \cref{tab:combined_ablation}(e), increasing $\alpha_t$ from $10^3$ to $10^4$ improves performance, but further increasing it to $10^5$ results in a decline. Consequently, we set $\alpha_t = 10^4$ for our experiments. Similarly, in \cref{tab:combined_ablation}(f), increasing $\tau$ to $15$ improves performance, while any further increase leads to a drop. Therefore, we set $\tau = 15$.

\subsection{Multi-Subject Customization}
To validate the effectiveness of our proposed \emph{test-time multi-subject fusion}, we compare \ours{} with state-of-the-art methods on the multi-subject customization task. Using the subject sets and prompts described in \cref{par:dataset}, we generate 720 videos and evaluate performance using \textit{CLIP-T}, \textit{CLIP-I}, \textit{DINO-I}, and \textit{T. Cons.}, following \cite{wang2024customvideo, chen2024disenstudio}. For fair comparison, we omit the additional bounding boxes required by DisenStudio. As shown in \cref{tab:multi_subject}, \ours{} outperforms the second-best method in \textit{CLIP-I}, \textit{DINO-I}, and \textit{T. Cons.}, and is comparable to CustomVideo in \textit{CLIP-T}.

\subsection{More Qualitative Results.}
In \cref{fig:supp_qualitative}, we present additional qualitative results of \ours{} customizing videos with multiple subjects and motion, successfully demonstrating diverse subject-motion combinations across various scenes, including cases with more than two subjects.

\end{document}